\documentclass[letterpaper]{article} 
\usepackage{aaai2026} 
\usepackage{times}  
\usepackage{helvet}  
\usepackage{courier}  
\usepackage[hyphens]{url}  
\usepackage{graphicx} 
\usepackage{natbib}  
\usepackage{caption} 
\usepackage{algorithm}
\usepackage{algorithmic}
\usepackage{amsmath}
\usepackage{multirow}
\usepackage{amssymb}

\usepackage{multirow}
\newcommand\boldp[1]{\vspace{.1cm}\noindent\textbf{#1.}\hspace{.065cm}}

\usepackage[table]{xcolor}
\definecolor{grey}{RGB}{128,128,128} 
\usepackage{subfigure}

\usepackage{newfloat}
\usepackage{listings}
\DeclareCaptionStyle{ruled}{labelfont=normalfont,labelsep=colon,strut=off} 
\floatstyle{ruled}
\newfloat{listing}{tb}{lst}{}
\floatname{listing}{Listing}
\nocopyright

\title{VDEGaussian: Video Diffusion Enhanced 4D Gaussian Splatting for Dynamic Urban Scenes Modeling}
\author{
    Bin Fan  \textsuperscript{\rm 1}\equalcontrib,
    Yuru Xiao \textsuperscript{\rm 1}\equalcontrib,
    Zihan Lin \textsuperscript{\rm 2}\equalcontrib,
    Chao Lu \textsuperscript{\rm 2},
    Deming Zhai \textsuperscript{\rm 1},
    Kui Jiang \textsuperscript{\rm 1},
    Wenbo Zhao \textsuperscript{\rm 1},
    Wei Zhang \textsuperscript{\rm 1},
    Junjun Jiang \textsuperscript{\rm 1},
    Huanran Wang \textsuperscript{\rm 2 \dag},
    Xianming Liu \textsuperscript{\rm 1}\thanks{Corresponding Authors}
}
\affiliations{

    \textsuperscript{\rm 1} Harbin Institute of Technology,
    \textsuperscript{\rm 2} Mach Drive
}

\usepackage{bibentry}

\begin{document}

\maketitle

\begin{abstract}
Dynamic urban scene modeling is a rapidly evolving area with broad applications. While current approaches leveraging neural radiance fields or Gaussian Splatting have achieved fine-grained reconstruction and high-fidelity novel view synthesis, they still face significant limitations. These often stem from a dependence on pre-calibrated object tracks or difficulties in accurately modeling fast-moving objects from undersampled capture, particularly due to challenges in handling temporal discontinuities.
To overcome these issues, we propose a novel video diffusion-enhanced 4D Gaussian Splatting framework. Our key insight is to distill robust, temporally consistent priors from a test-time adapted video diffusion model. To ensure precise pose alignment and effective integration of this denoised content, we introduce two core innovations: a joint timestamp optimization strategy that refines interpolated frame poses, and an uncertainty distillation method that adaptively extracts target content while preserving well-reconstructed regions. Extensive experiments demonstrate that our method significantly enhances dynamic modeling, especially for fast-moving objects, achieving an approximate PSNR gain of 2 dB for novel view synthesis over baseline approaches. (Project Page: https://pulangk97.github.io/VDEGaussian-Project/)

\end{abstract}

\section{Introduction}
Dynamic urban scene modeling is a critical task with broad applications in areas such as scene simulation, autonomous driving, and virtual reality \cite{yang2023learning, yan2024street, xie2023snerf, guo2023streetsurf, kulhanek2024wildgaussians, fischer2024dynamic, zhou2024hugs}. A key challenge within this domain lies in accurately modeling dynamic actors to ensure high-fidelity view synthesis at novel timestamps—a capability essential for continuous scene modeling and advanced simulation environments.

\begin{figure}
    \centering
    \includegraphics[width=1\linewidth]{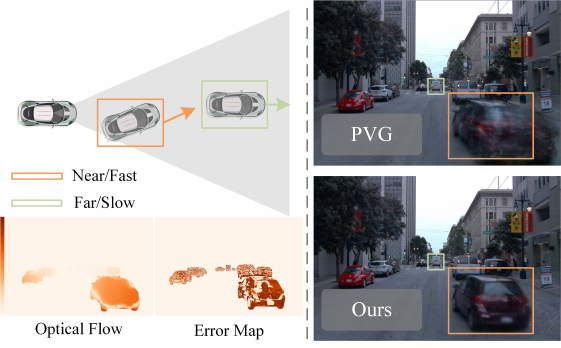}
    \caption{Motivation. We analyze the optical flow magnitude between adjacent frames and the error map of the rendered mid-frame. Our results demonstrate a strong correlation between flow magnitude and dynamic modeling accuracy, highlighting temporal discontinuity as a key challenge.}
    \label{fig:motivation}
\end{figure}

Recent advances based on Neural Radiance Fields (NeRF) \cite{mildenhall2021nerf} or 3D Gaussian Splatting (3DGS) \cite{kerbl20233d} often leverage either scene graphs \cite{ost2021neural, yan2024street, zhou2024drivinggaussian, chen2025omnire, fischer2024dynamic} or time-dependent fields \cite{yangemernerf, huang2024sgaussian, chen2023periodic, nguyen2024rodus}. Scene-graph-based methods, for instance, rely on calibrated 3D bounding boxes to localize dynamic actors. This dependency makes them susceptible to artifacts like shifts or deformations during novel view synthesis, especially when these boxes are noisy or misaligned.
To address these limitations, self-supervised scene decomposition techniques \cite{yangemernerf, huang2024sgaussian} equipped with time-dependent fields have gained momentum, enabling temporal modeling without the need for annotated 3D boxes. Periodic Vibration Gaussian (PVG) \cite{chen2023periodic} further streamlines decomposition by incorporating lifespan mechanisms. Building on PVG, DeSiRe-GS \cite{peng2024desiregs4dstreetgaussians} introduces an additional dynamic mask prior and a velocity map correlation strategy to enhance dynamic decomposition.
Despite these improvements, accurately modeling dynamic objects without calibrated tracks remains a significant hurdle. Artifacts frequently appear in synthesized views at novel timestamps, particularly for nearby or fast-moving objects exhibiting substantial optical flow (see Fig. \ref{fig:motivation}). These issues primarily stem from undersampled dynamic content and the inherent limitations of self-supervised learning, which collectively impede the accurate reconstruction of continuous temporal evolution.

In this paper, we investigate the root cause of unreliable dynamic modeling, identifying it as the temporal discontinuity of input images. As illustrated in Fig.~\ref{fig:motivation}, we compute the optical flow between adjacent input frames and the error map of the corresponding novel mid-frame. Our results show a clear correlation: larger optical flow—typically arising from nearby or fast-moving vehicles—correlates with higher errors in the novel mid-frame.
To address this issue, we propose a novel method that leverages temporal consistency priors recently introduced in video diffusion models. While a straightforward approach would be to generate pseudo-images for interpolated frames, current video diffusion models lack robust pixel-level 3D consistency. This limitation makes it difficult to directly use these interpolated images to improve dynamic modeling, as it introduces ambiguity and conflicts with the inherent 3D consistency of 3DGS. Furthermore, the absence of fine-grained pose control complicates the alignment between interpolated views and the generated frames.
To overcome these challenges, we introduce an uncertainty distillation strategy for target content extraction alongside a joint timestamp optimization strategy for pose alignment. These components consistently distill temporal consistency priors from the video diffusion model to specifically address distorted objects while maintaining well-reconstructed areas undisturbed.
The main contributions of this paper are summarized as follows:

\begin{itemize}
    \item We propose a novel framework for enhancing dynamic modeling in urban scenes based on 4DGS, tailored to address challenges arising from temporal discontinuity.

    \item We introduce an uncertainty distillation strategy and a joint timestamp optimization strategy to accurately extract desired content from the video generation model while ensuring aligned poses.

    \item Our framework significantly enhances self-supervised dynamic urban scene modeling, achieving nearly a 2dB PSNR improvement and delivering higher visual quality without additional data capture or calibration.
\end{itemize}

\section{Related Work}
\boldp{Dynamic Urban Scene Reconstruction} Dynamic scene reconstruction using time-dependent Neural Radiance Fields (NeRF) \cite{mildenhall2021nerf} or 3D Gaussian Splatting (3DGS) \cite{kerbl20233d} has advanced significantly in recent years due to their ability to render highly realistic novel views. Deformation-based approaches \cite{park2021nerfies, pumarola2021d, yang2024deformable, cao2023hexplane, Wu_2024_CVPR}, such as D-NeRF \cite{pumarola2021d} and DeformableGS \cite{yang2024deformable}, employ timestamp-conditioned neural networks to model deformation fields. By leveraging the implicit prior of these networks, they achieve temporally continuous dynamic modeling. Recent work also explores explicit representations of dynamic 3DGS through Gaussian interpolation strategies \cite{lee2024fully}. However, these methods typically struggle with large-scale dynamic urban scenes.

Alternatively, Neural Scene Graphs (NSG) \cite{ost2021neural} represent urban scenes using graph structures, where each node corresponds to an independent implicit network. Similarly, S-NeRF \cite{xie2023snerf} decouples static and dynamic components, placing moving objects via 3D detection algorithms. DrivingGaussian \cite{zhou2024drivinggaussian} combines NSG with incremental reconstruction to address inefficiencies caused by excessive Gaussian primitives, while OmniRe \cite{chen2025omnire} enhances pedestrian and cyclist modeling using auxiliary deformation and SMPL nodes. Although NSG-based methods enable scene editing for simulation, their heavy dependency on calibrated 3D bounding boxes limits their robustness. Techniques like VEGS \cite{hwang2024vegs} jointly optimize bounding boxes, yet their accuracy remains constrained.

To reduce reliance on ground-truth tracks \cite{peng2024desiregs4dstreetgaussians, sun2024splatflow, yangemernerf, huang2024sgaussian, chen2023periodic}, EmerNeRF \cite{yangemernerf} introduces a self-supervised scene decomposition method for dynamic scenes modeling. S3Gaussian \cite{huang2024sgaussian} adapts Hexplane \cite{cao2023hexplane} to street scenes by separating time-independent and time-dependent components into distinct tri-plane encoders. However, this approach requires extended training and an additional initialization stage. Periodic Vibration Gaussian (PVG) \cite{chen2023periodic} addresses this by proposing a lifespan mechanism for the rapid decomposition of static and dynamic Gaussian primitives.
Despite avoiding ground-truth dependencies, these methods still face challenges: they tend to overfit observed timestamps and exhibit poor performance on intermediate frames, particularly for fast-moving objects with large optical flow (see Fig. \ref{fig:motivation}). This limitation suggests unreliable dynamic learning when object observations are discontinuous. In this paper, we address the time-domain undersampling problem and propose improvements to dynamic modeling for self-supervised urban scene reconstruction.

\boldp{Video Generation as Prior for Scene Reconstruction}
Video generation \cite{xing2023dynamicrafter, he2022lvdm, chen2024videocrafter2, ma2025latte, yang2024cogvideox, blattmann2023stablevideodiffusionscaling, yu2024viewcraftertamingvideodiffusion} has advanced significantly, leveraging its inherent world prior to adapt reconstruction models for addressing ill-posed problems or enhancing performance. For instance, 3DGS-Enhancer \cite{liu20243dgs} employs a video diffusion model to resolve sparse-view issues by generating dense interpolated pseudo-views, while ViewExtrapolator \cite{liu2024novel} achieves view extrapolation using a pre-trained video diffusion model without additional fine-tuning. Similarly, StreetCrafter \cite{yan2024streetcrafter} utilizes a LiDAR-conditioned video diffusion model for novel trajectory generation, improving scene reconstruction completeness, and ReconDreamer \cite{ni2024recondreamer, zhao2025recondreamer++} extrapolates trajectories by restoring degraded rendered videos from novel viewpoints. However, none of these methods leverage the inherent temporal consistency of video diffusion models to tackle temporal discontinuities in dynamic reconstruction.

\section{Method}
\subsection{Preliminaries}
\boldp{3D Gaussian Splatting}
3D Gaussian Splatting (3DGS) represents a scene using a collection of 3D Gaussian primitives. Each primitive is defined by three differentiable parameters: a position vector  $\boldsymbol{\mu} \in \mathbb{R}^3$, a rotation quaternion $\mathbf{q} \in \mathbb{R}^4$, and a scaling vector $\mathbf{s} \in \mathbb{R}^3$:

\begin{equation}
    G(\mathbf{x}) = e^{-\frac{1}{2}(\mathbf{x}-\boldsymbol{\mu})^T{\Sigma}^{-1}(\mathbf{x}-\boldsymbol{\mu})},
\end{equation}
where the covariance matrix ${\Sigma} \in \mathbb{R}^{3\times3}$ is computed using the scaling vector $\mathbf{s}$ and the rotation quaternion $\mathbf{q}$. The Gaussian primitives are then projected onto the imaging plane. The covariance matrix on the imaging plane is derived as:
\begin{equation}
    {\Sigma}' = JW{\Sigma}W^{T}J^{T},
\end{equation}
where $W$ is the view transformation matrix and $J$ is the Jacobian of the affine approximation of the projective transformation \cite{zwicker2001ewa}. 3DGS employs volume rendering to composite the 2D Gaussians in depth order:
\begin{equation}
    \mathbf{c} = \sum_{i\in N}\mathbf{c}_i\hat{\alpha}_i \prod_{j=1}^{i-1}(1-\hat{\alpha}_j),
\end{equation}
where $\mathbf{c}_i$ is the color of the $i$-th Gaussian $G_i(\mathbf{x})$, derived from spherical harmonic coefficients, and $\hat{\alpha}_i$ is computed from the projected 2D Gaussian and its opacity $o$.

\boldp{Periodic Vibration Gaussian} Periodic Vibration Gaussian (PVG) incorporates an additional velocity vector $v$ to model the oscillatory motion of 3D Gaussians (Eq. \ref{eq:mu}). It also introduces a life-span parameter $\beta$ to modulate opacity over time (Eq. \ref{eq:op}), enabling efficient decomposition of static and dynamic Gaussians. The dynamic Gaussians are modeled as:
\begin{equation}
    \boldsymbol{\widetilde\mu}(t) = \boldsymbol{\mu} + \frac{l}{2\pi}\cdot sin(\frac{2\pi(t-\tau)}{l})\cdot \mathbf{v},
    \label{eq:mu}
\end{equation}

\begin{equation}
    \widetilde o(t) = o \cdot e^{-\frac{1}{2}(t-\tau)^2\beta^{-2}},
    \label{eq:op}
\end{equation}
where $\boldsymbol{\widetilde\mu}$ represents the oscillatory motion centered at $\boldsymbol{\mu}$, peaking at time $\tau$,  while $\widetilde o(t)$ denotes the vibrating opacity that decays from its peak value with a rate governed by the life-span parameter $\beta$. The hyper-parameter $l$, defining the cycle length as a scene prior.

\subsection{Overview}
\begin{figure*}
    \centering
    \includegraphics[width=1\linewidth]{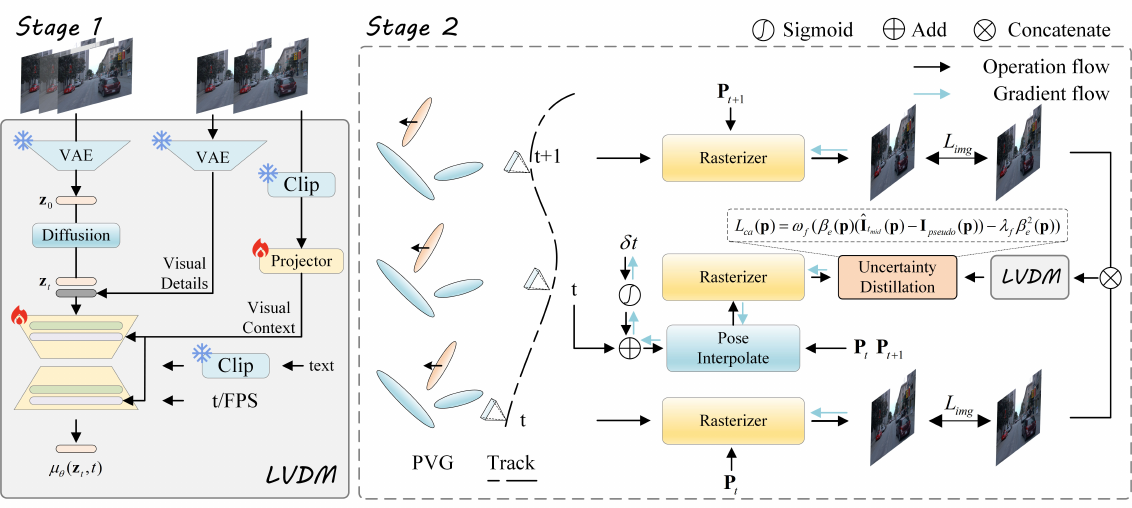}
    \caption{Our method employs a two-stage training approach. Initially, we fine-tune a latent video diffusion model to the target scene and desired frame count. Subsequently, we perform 4D reconstruction, utilizing the adapted model as a mid-frame prior.}
    \label{fig:method}
\end{figure*}

As illustrated in Fig. \ref{fig:method}, our approach employs a two-stage architecture. The first stage adapts a pre-trained video generation model to flexibly produce high-quality interpolated frames with variable frame counts. The adapted model then feeds into the second stage, where it enhances time-domain continuity by providing intermediate frame priors.
We represent dynamic scenes using Periodic Vibration Gaussian (PVG) within a 4DGS framework. However, directly incorporating generated frames as pseudo-views introduces pixel-level misalignment and 3D inconsistencies between the interpolated rendered views and the generated frames. To mitigate this, we design an uncertainty distillation strategy and jointly optimize the interpolated pose, restricting the optimization space to the challenging distorted areas.

\begin{figure}[!htp]
    \centering
    \includegraphics[width=1\linewidth]{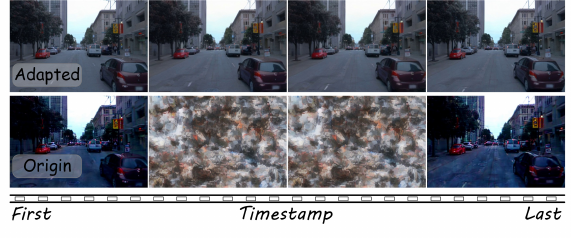}
    \caption{The base video generation model lacks flexibility in the number of generated frames. However, our test-time adaptation effectively resolves this limitation.}
    \label{fig:video_adapt}
\end{figure}

\subsection{Scene Adaptation}
Video generation models have been extensively studied for their world modeling capabilities and temporal consistency. However, despite being pre-trained on large-scale video datasets, their learned generation distribution often misaligns with autonomous driving data, which involves complex real-world scenes and high-speed motion. Additionally, existing methods typically generate a fixed number of frames, limiting their adaptability to variable-length video synthesis.
In our framework, excessive frame generation exacerbates 3D inconsistencies in the second stage and undermines alignment optimization strategies. Moreover, producing redundant frames reduces efficiency for downstream reconstruction tasks. To address this, we fine-tune the video generation model to adapt to specific scenes and variable frame counts. As illustrated in Fig. \ref{fig:video_adapt}, we reduce the number of interpolated frames from 16 to 4. The original model fails to produce accurate videos under this setting, but low-cost test-time fine-tuning successfully adapts it to the target scene and configuration.

In this paper, we build our model upon DynamiCrafter \cite{xing2023dynamicrafter}. To adapt the generation model to a specific scene while preserving the inherent temporal consistency, we introduce learnable low-rank adaptation (LoRA \cite{hu2022lora}) layers to the frozen spatial attention and temporal attention layers. Additionally, we find that jointly fine-tuning the visual context injection module improves alignment between the visual context of autonomous driving scenes and the learned latent-space concepts, leading to better performance.

\subsection{Joint Timestamp Optimization}
One challenge in leveraging video generation priors is ensuring pose alignment between rendered interpolated images and generated frames. Current methods adapt video generation models using camera pose control \cite{liu20243dgs}, projected LiDAR images \cite{yan2024streetcrafter}, or rendered degraded Gaussian images \cite{ni2024recondreamer}. However, integrating additional control signals often requires costly fine-tuning and may compromise inherent diversity.
In this paper, we propose a novel approach to actively align the pose of rendered interpolated images with generated frames. Our key idea is to jointly optimize the interpolated pose to find the rendered image most similar to the generated results. Specifically, we compute the interpolated pose from two nearest poses, $\mathbf{P}_t=(\mathbf{R}_t,\mathbf{T}_t)$ and $\mathbf{P}_{t+1}=(\mathbf{R}_{t+1}, \mathbf{T}_{t+1})$, where $t$ and $t+1$ denote frame indices. The rotation matrices $\mathbf{R}_t$ and $\mathbf{R}_{t+1}$ can also be represented as quaternions $\mathbf{q}_t$ and $\mathbf{q}_{t+1}$ for spherical linear interpolation. Given $\mathbf{q}_t$, $\mathbf{q}_{t+1}$, and a mid-indice $t_{mid}$, the interpolated quaternion is computed as:

\begin{equation}
    q_{mid} = \frac{q_t sin((1-t_{mid})\theta)+q_{t+1} sin(t_{mid}\theta)}{sin(\theta)},
    \label{eq:q_interp}
\end{equation}
with
\begin{equation}
    \theta = arccos(q_t \cdot q_{t+1}).
\end{equation}
Since the camera's capture interval between  $t+1$ and $t$ is very short, the quaternions change gradually. As a result, the dot product of the two quaternions remains close to 1, allowing Eq. \ref{eq:q_interp} to be approximated and simplified.
\begin{equation}
    q_{mid} \approx (1-t_{mid})q_{t}+t_{mid}q_{t+1}.
    \label{eq:sim_q_interp}
\end{equation}
We employ the simplified representation in Eq. \ref{eq:sim_q_interp} for pose interpolation, as it is differentiable and computationally efficient. To ensure stable convergence, we initialize the optimization of $t_{mid}$ from the nearest plausible point and constrain its search range. This is achieved by adding a learnable bias term $\delta t$, activated by a sigmoid function to restrict deviations, to the start indice $t$.
\begin{equation}
    t_{mid} = t + \sigma(\delta t),
\end{equation}
where $\sigma$ represents the sigmoid function and the learnable bias $\delta t$ is initialized to zero. The transmittance parameters $\mathbf{T}_t$ and $\mathbf{T}_{t+1}$ are likewise amenable to interpolation as
\begin{equation}
    \mathbf{T}_{mid} = (1-t_{mid})\mathbf{T}_t + t_{mid} \mathbf{T}_{t+1} .
\end{equation}
The interpolated camera pose parameters are passed into a rasterization module that supports camera pose backpropagation. This enables gradient flow from the image loss to the learnable bias $\delta t$.


\subsection{Uncertainty Distillation.}
A significant challenge in integrating generated content into scene reconstruction lies in the inherent 3D inconsistency present within generated images. To effectively address this, particularly for fast-moving objects, and to mitigate the adverse impact of this 3D inconsistency on well-reconstructed static backgrounds or slow-moving scene elements, we propose an uncertainty distillation method. Inspired by recent advances in distractor-free NeRF approaches \cite{ren2024nerf, martin2021nerf}, our method adaptively extracts reliable information from generated images. This is achieved through the following regularization formula:

\begin{equation}
    \label{eq:lca}
    L_{ca}(\mathbf{p}) = \omega_{f}(\beta_e(\mathbf{p})(\hat{\mathbf{I}}_{t_{mid}}(\mathbf{p}) - \mathbf{I}_{pseudo}(\mathbf{p}))^2-\lambda_f\beta_e^2(\mathbf{p})),
\end{equation}
where $\beta_e$ is a learnable uncertainty map for each pseudo-view. Here, $\mathbf{p}$ denotes a single pixel, $\mathbf{I}_{pseudo}$ represents the image generated by the adapted video diffusion model, and $\hat{\mathbf{I}}_{t_{mid}}$ is the rendered view at the novel timestamp $t_{mid}$, downsampled to match the resolution of the generated image. To enhance the understanding of the uncertainty map's effect, we establish a mathematical correspondence between the optimized uncertainty map and the loss function. We begin by deriving the partial derivative with respect to $\beta_e(\mathbf{p})$:
\begin{equation}
    \frac{dL_{ca}(\mathbf{p})}{d\beta_e(\mathbf{p})} = \omega_f((\hat{\mathbf{I}}_{t_{mid}}(\mathbf{p}) - \mathbf{I}_{pseudo}(\mathbf{p}))^2 - 2\lambda_f\beta_e(\mathbf{p})).
\end{equation}
Setting the derivative function to zero yields the optimized solution for $\beta_e(\mathbf{p})$ as
\begin{equation}
\label{eq:ucs}
   \frac{dL_{ca}(\mathbf{p})}{d\beta_e(\mathbf{p})}=0 \Rightarrow\beta_e(\mathbf{p}) = \frac{(\hat{\mathbf{I}}_{t_{mid}}(\mathbf{p}) - \mathbf{I}_{pseudo}(\mathbf{p}))^2}{2\lambda_f}.
\end{equation}
According to Eq. \ref{eq:ucs}, the optimized $\beta_e$ is positively correlated with the error map between the rendered novel view and the generated pseudo view. This formulation enhances its influence on regions with large errors (e.g., fast-moving objects) while reducing the impact on well-reconstructed areas with smaller errors, thereby mitigating disturbances in the background or on slow-moving objects. To alleviate overfitting and the detrimental effect of high-frequency noise in the uncertainty map, we apply Total Variation (TV) normalization to the uncertainty map during training:
\begin{equation}
    L_{tv} =  \omega_{tv}\text{TV}(\beta_e).
\end{equation}

\section{Experiments}
\subsection{Experimental Setup}
\boldp{Datasets \& Metrics}
To rigorously evaluate our method, we established a small benchmark for self-supervised dynamic urban scene reconstruction by selecting six challenging dynamic scenes from the Waymo Open Dataset \cite{sun2020scalability} (segment IDs: 102319, 103913, 106250, 109636, 121618, and 225932). For this benchmark, we trained on the first 50 frames of the front camera, sampling test images every four views. Additionally, for a comprehensive evaluation, we tested our method on NOTR's dynamic32 dataset \cite{yangemernerf}, utilizing all training views. For quantitative assessment of novel view synthesis, we report PSNR, SSIM, and LPIPS metrics.

\boldp{Baselines}
We evaluate our method against the PVG codebase \cite{chen2023periodic} and state-of-the-art self-supervised approaches, including EmerNeRF \cite{yangemernerf}, 3DGS-based S3Gaussian\cite{huang2024sgaussian}, and the recent DeSiReGS \cite{peng2024desiregs4dstreetgaussians}. Unlike these methods, scene graph-based approaches \cite{yan2024street, zhou2024drivinggaussian, chen2025omnire} do not learn dynamics directly from input videos in a self-supervised manner, thus avoiding the issue shown in Fig. \ref{fig:motivation}. Since such comparisons would be uninformative, we exclude these methods from our analysis.

\boldp{Implementation Details}
Our method is implemented based on PVG \cite{chen2023periodic}, a self-supervised architecture for dynamic urban scene reconstruction. To enable gradient backpropagation of camera poses in the Gaussian rasterizer, we integrate the back propagation equation from iComMa \cite{sun2023icomma} into the PVG framework. For frame interpolation, we use a test-time-adapted latent video diffusion model (DynamiCrafter \cite{xing2023dynamicrafter}), fine-tuned on 3-frame video clips extracted from the input training views of the test scene. The dataset consists of nearly 50 clips.
We introduce learnable LoRA adapters to the spatial attention layers and temporal attention layers. These adapters are jointly optimized with DynamiCrafter’s visual context injection module. We fine-tune these modules on the training clips for 1,000 iterations, maintaining all other original training hyperparameters. The adapted model then advances to the second stage for intermediate image generation. In the second stage, we adopt PVG’s progressive training strategy, incrementally increasing the resolution of Gaussians from low (16× downsampled) to high (2× downsampled) resolution. Uncertainty distillation, as shown in Eq. \ref{eq:lca}, is performed every four iterations to ensure balanced training for each input view. In all our experiments, we set the hyperparameters $\omega_f$ , $\omega_{tv}$ and $\lambda_f$ to 1, 0.001, and 1, respectively. We use the Adam \cite{kingma2014adam} to optimize the learnable bias $\delta t$ and the uncertainty map $\beta_e$, with a learning rate of 0.01.

\subsection{Comparisons with the baselines}

\begin{figure*}
    \centering
    
    \includegraphics[width=0.8\linewidth]{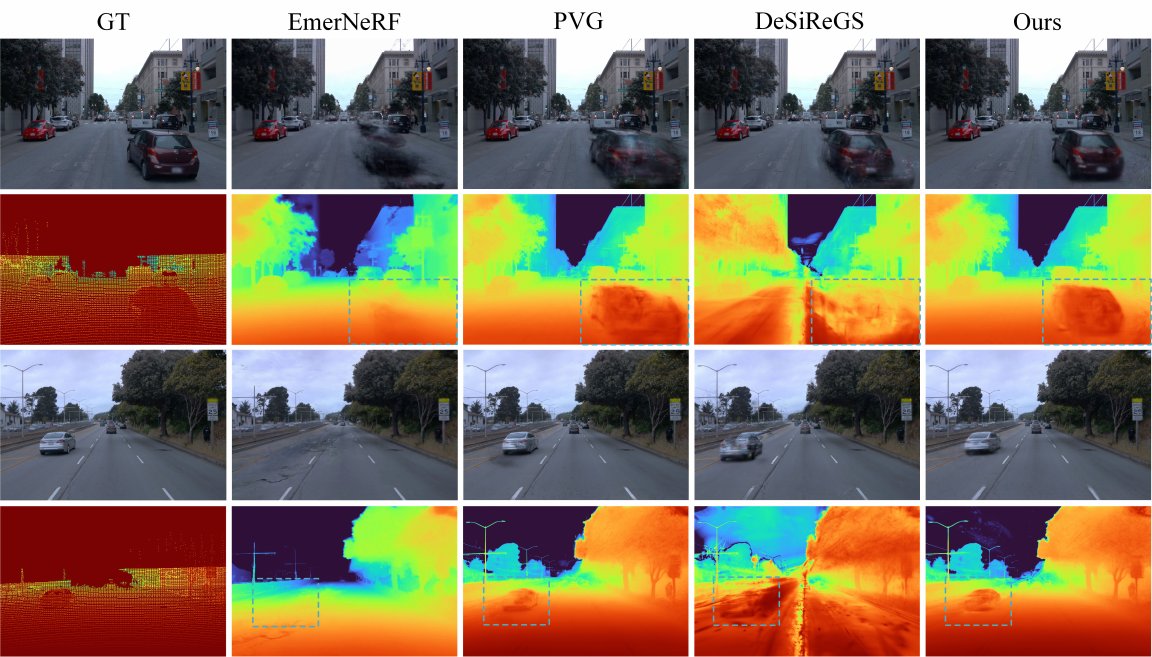}
    \caption{Qualitative comparisons of novel-view rendering results on the Waymo Open Dataset demonstrate that our method outperforms baselines, generating sharper outputs with fewer artifacts and distortions.}
    \label{fig:waymo_high}
\end{figure*}

\boldp{Evaluation on the Waymo Dataset}
We initially evaluate our method on a subset of Waymo, specifically 6 challenging scenes containing similar bad cases as shown in Fig. \ref{fig:motivation}. Qualitative and quantitative results are presented in Fig. \ref{fig:waymo_high} and Tab. \ref{tab:subset}, respectively. To validate dynamic learning for high-speed motion, we assess rendered views at novel timestamps.
Our approach distills temporal priors for low-resolution rendered images, effectively capturing motion dynamics and enhancing detail refinement. The resulting novel views exhibit fewer artifacts and greater 3D consistency. Quantitatively, our method outperforms the baseline PVG and the more recent method, DeSiReGS, by 2 dB in PSNR. For a thorough evaluation, we present the quantitative evaluation results on the Waymo NOTR dynamic32 dataset in Tab. \ref{tab:waymo}. These quantitative results indicate that our method achieves comprehensive improvements compared with the baseline PVG, further verifying its robustness across various scenes, even without challenging events.

\begin{table}[!htp]
    \centering
    \small
    \setlength{\tabcolsep}{1mm} 
    {
    \begin{tabular}{c|ccc} \hline
         & PSNR$\uparrow$ & SSIM$\uparrow$ & LPIPS$\downarrow$  \\ \hline
        EmerNeRF & 26.89 & 0.831 & 0.332  \\
        S3Gaussian &  22.62 & 0.831 & {0.230}  \\
        PVG & {28.31} & {0.884} & {0.213}  \\
        DeSiReGS & {28.16} & {0.880} & {0.219}  \\
        Ours & \textbf{30.49} & \textbf{0.894} & \textbf{0.209} \\\hline
    \end{tabular}}
    \caption{Quantitative Results on a Subset of Waymo.}
    \label{tab:subset}
\end{table}

To evaluate the effectiveness of our method in learning scene dynamics, we compare the rendered novel-view depth results with baseline methods and ground-truth LiDAR depth images (Fig. \ref{fig:waymo_high}). Our approach, which incorporates temporal priors distilled from a video generation model, demonstrates superior geometric consistency at novel timestamps. In contrast, the baseline PVG exhibits significant artifacts, particularly around moving objects, such as distortions behind the red car (second row) and the white car (third row). Our method effectively mitigates these issues, producing more accurate depth rendering. The improved placement of Gaussians at novel timestamps further confirms the enhanced dynamic modeling capability of our framework.

\begin{table}[!htp]
    \centering
    \small
    \setlength{\tabcolsep}{1mm} 
    {
    \begin{tabular}{c|ccc} \hline
         &PSNR$\uparrow$ & SSIM$\uparrow$ & LPIPS$\downarrow$  \\ \hline
        S3Gaussian & {27.76} & {0.877} & \textbf{0.159}  \\
        PVG & {28.73} & {0.874} & {0.245}  \\
        Ours & \textbf{29.33} & \textbf{0.879} & {0.239} \\\hline
        \end{tabular}}
    \caption{Quantitative Results on the Waymo NOTR Dataset.}
    \label{tab:waymo}
\end{table}

\subsection{Method Analysis}
\begin{figure}[!htp]
    \centering
    \includegraphics[width=0.85\linewidth]{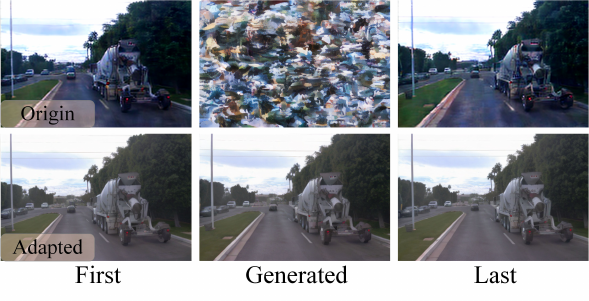}
    \caption{Comparison of Original and Adapted Video Generation Models.}
    \label{fig:adaptation}
\end{figure}

\boldp{Analysis of Test Time Adaptation} 
Video generation models learn data distributions for a fixed number of frames, often failing when adjusted to different video lengths. A model capable of generating a flexible number of frames, however, could adapt more efficiently to specific applications.
In this work, we demonstrate that test-time adaptation effectively addresses this issue without costly retraining (Fig. \ref{fig:adaptation}). Experiments varying the frame count from 16 to 3 reveal that the original model produces incorrect frames with distorted color tones, whereas a lightweight test-time adaptation—applied to a small dataset—surprisingly restores temporal consistency and visual quality (Fig. \ref{fig:adaptation}).
Notably, while the base model was fine-tuned on a 3-frame setting, the adapted model generalizes to other lengths (e.g., 4 frames), as shown in Fig. \ref{fig:video_adapt}. This flexibility enhances its applicability across diverse scenarios.

\begin{figure*}
    \centering
    \subfigure[Scene 16]{
        \includegraphics[width=0.23\textwidth]{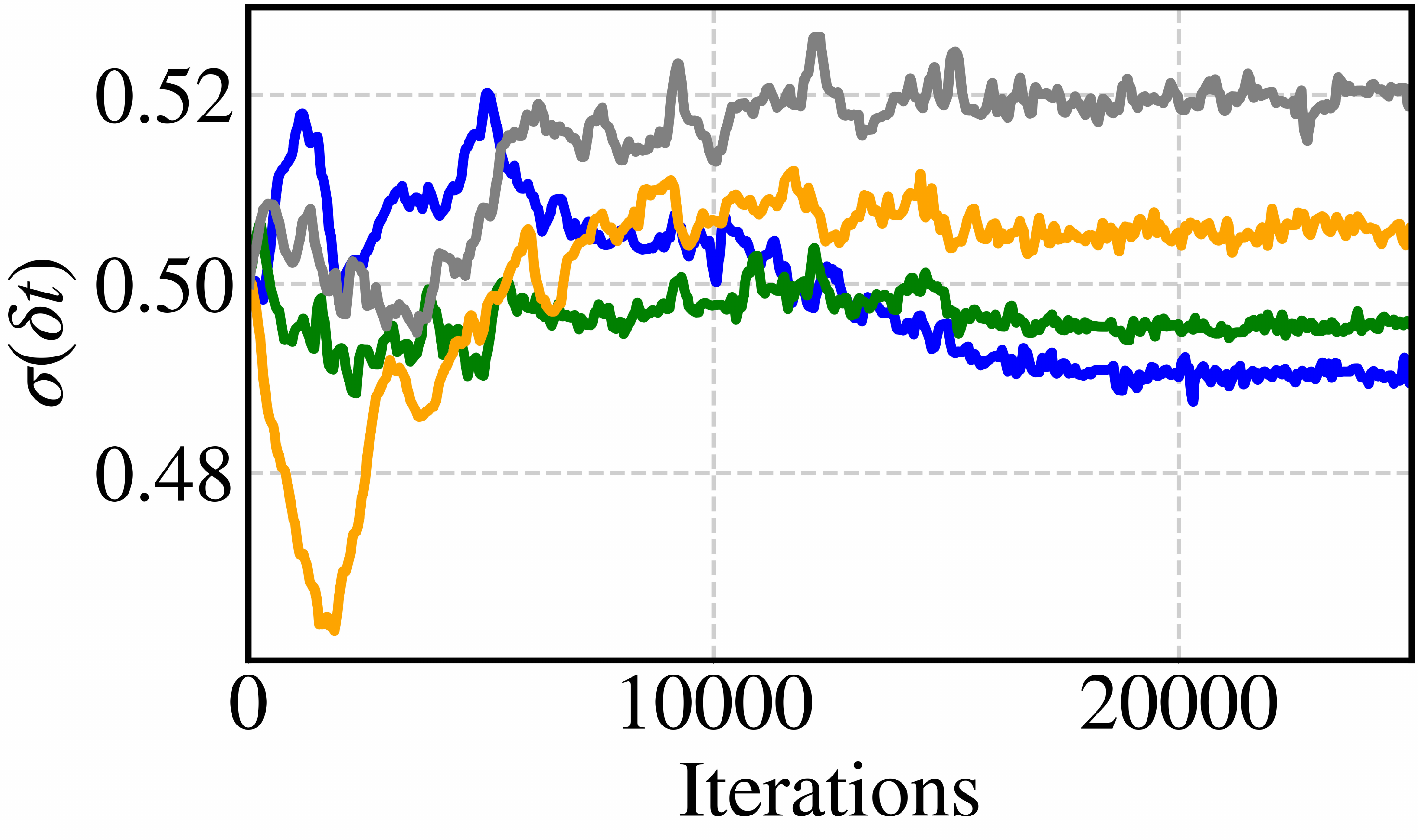}
        \label{fig:sub1}
    }
    \hfill
    \subfigure[Scene 34]{
        \includegraphics[width=0.23\textwidth]{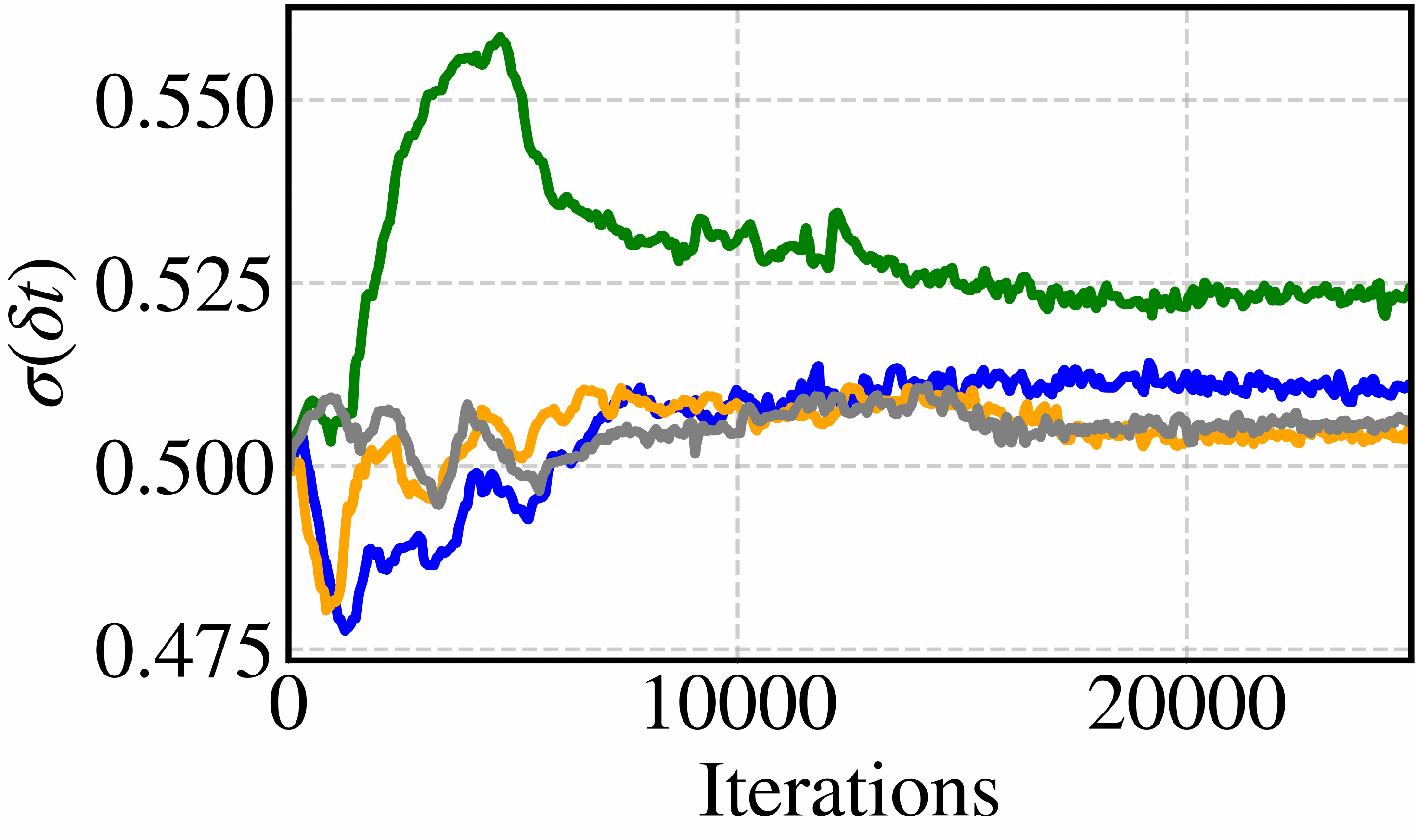}
        \label{fig:sub2}
    }
    \hfill
    \subfigure[Scene 96]{
        \includegraphics[width=0.23\textwidth]{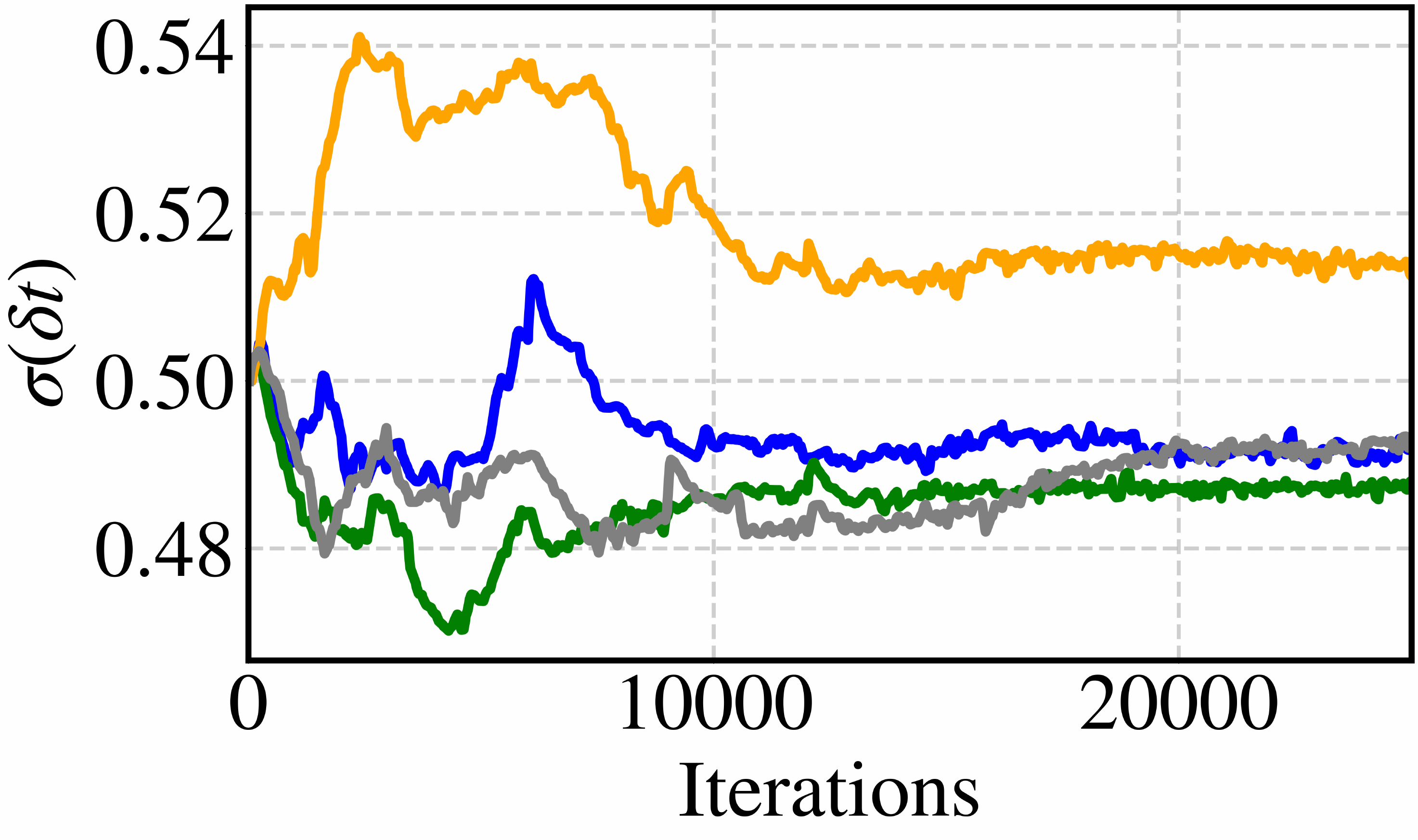}
        \label{fig:sub3}
    }
    \hfill
    \subfigure[$\delta (t)$ for frame 0 in scene 16]{
        \includegraphics[width=0.23\textwidth]{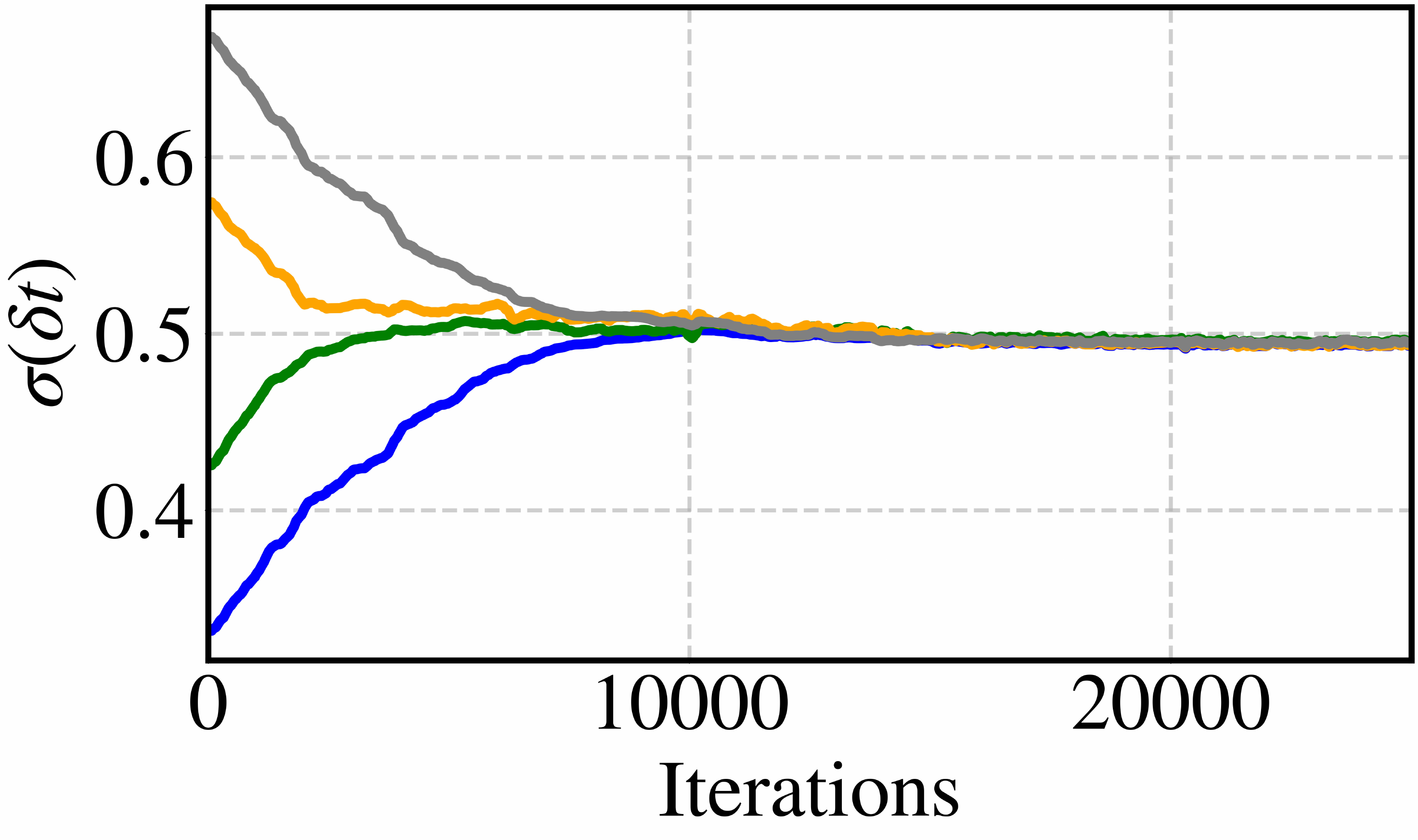}
        \label{fig:sub4}
    }
    
    \caption{We conduct experiments to evaluate the optimization process of the learnable $\delta t$. In subfigures (a), (b), and (c), we present the optimization curves for different test frames (IDs 0, 3, 6, and 9), with each frame represented by a distinct color—blue, green, orange, and gray, respectively. In subfigure (d), we analyze the optimization trajectory across varying initial values of $\delta t$ to verify the stability of the pose optimization. Here, each color corresponds to a different initial value.}
    \label{fig:time_stamp_optimization}
\end{figure*}

\boldp{Analysis of Timestamp Optimization}
To better integrate generated content into 3D scenes, we propose a timestamp optimization strategy that aligns the interpolated rendered images with the generated images. Instead of optimizing camera poses directly, we constrain the optimization to a single parameter, $\delta t$ , simplifying the objective function and improving convergence.
As shown in Fig. \ref{fig:time_stamp_optimization},   we extract the activated learnable bias $\sigma (\delta t)$ at each training step and plot its trajectory. The resulting curve (Fig. \ref{fig:time_stamp_optimization} (a), (b), (c)) demonstrates that $\sigma (\delta t)$—which governs the interpolated pose—converges stably during joint optimization. 
However, the curves also reveal that the generated frame’s pose is not perfectly aligned with the midpoint of adjacent poses. This discrepancy may stem from inherent biases learned by the video generation model during large-scale pretraining. Addressing this bias in pretraining could be a promising direction for downstream tasks.
We further assess the optimization of  $\delta (t)$  across different initial values (Fig. \ref{fig:time_stamp_optimization} (d)). The results demonstrate that our method is robust to initialization, consistently converging to similar values even when starting far from the optimum. The rapid and stable convergence demonstrates our method’s effectiveness in aligning interpolated frames with generated frames.

\boldp{Analysis of Uncertainty Distillation} 
\begin{figure}
    \centering
    \includegraphics[width=0.9\linewidth]{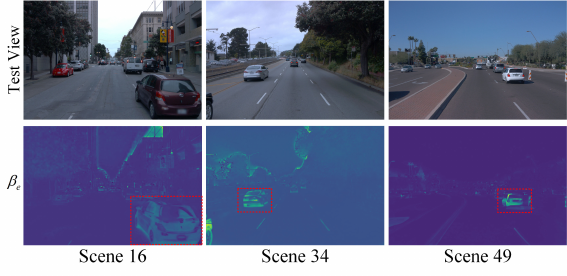}
    \caption{We present the visualization of the uncertainty map. The learnable uncertainty map successfully identifies fast-moving objects with high weights and assigns low weights to the static background.}
    \label{fig:UE}
\end{figure}
Theoretically, as formalized in Eq. \ref{eq:ucs}, the optimized uncertainty map can identify both poorly reconstructed areas (e.g., fast-moving objects) and well-reconstructed areas (e.g., background and slow-moving objects). We visualize the uncertainty map $\beta_e$ in Fig. \ref{fig:UE} to validate this theory. During the joint optimization of the uncertainty map and the Gaussians, the map successfully assigns higher values to poorly reconstructed dynamic regions and lower values to well-reconstructed static backgrounds, thereby reducing the influence of the generated image on these respective areas.

\begin{figure}[!htp]
    \centering
    \includegraphics[width=1\linewidth]{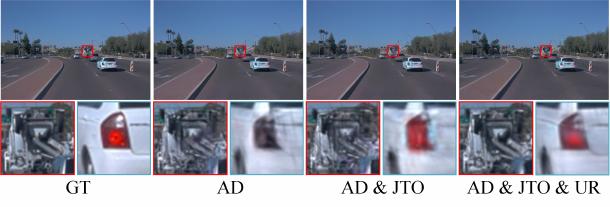}
    \caption{Qualitative Ablation Study on the Waymo Dataset.}
    \label{fig:abl}
\end{figure}

\boldp{Ablation Study}
To thoroughly evaluate the contribution of each component in our method, we conduct an ablation study by decomposing the full pipeline into three key parts: (1) test-time adaptation (AD), (2) joint timestamp optimization (JTO), and (3) the uncertainty distillation (UD) strategy. We systematically integrate each component, enabling a direct assessment of their individual effectiveness. The qualitative and quantitative assessments are presented in Fig.~\ref{fig:abl} and Tab.~\ref{tab:abl}, respectively. The quantitative results demonstrate that each component significantly contributes to our full method.
Specifically, without test-time adaptation, the generation process for video length 3 failed, leading to extreme degradation of the original performance. In contrast, integrating pseudo images from the adapted model achieves slightly better performance compared with the baseline PVG. Furthermore, the joint timestamp optimization, designed for pose alignment, yields slight improvements in both quantitative and qualitative results.

Notably, the uncertainty distillation strategy shows a significant effect, improving PSNR by nearly 2 dB compared to our method without implementing the uncertainty map (i.e., directly applying pseudo loss on interpolated images). The qualitative results further verify this effect. Without the uncertainty map, the truck positioned far away (shown in the red box in Fig. \ref{fig:abl}) is significantly distorted and blurred, whereas our uncertainty distillation successfully addresses this issue, achieving more realistic rendering for both near and far moving objects.

\begin{table}[!htp]
    \centering 

    \small
    \begin{tabular}{c|ccc|ccc} 
    \hline

    & AD & JTO & UD & PSNR$\uparrow$ & SSIM$\uparrow$ & LPIPS$\downarrow$  \\ \hline
  PVG & $\times$  & $\times$ &$\times$ &  28.31 & 0.884   & 0.213   \\ \hline
       \multirow{3}{*}{Ours}      & $\times$  & $\times$ & $\times$ & 9.65  & 0.339  & 0.649  \\        
                        & $\checkmark$  & $\times$ & $\times$ & 28.73  & 0.872  & 0.226  \\
                      & $\checkmark$  & $\checkmark$ & $\times$  & 28.77 & 0.873 & 0.224 \\ 
                     & $\checkmark$  & $\checkmark$ &  $\checkmark$  & 30.49 & 0.892 & 0.209 \\ \hline
    \end{tabular}
    \caption{Quantitative Ablation Study on the Waymo Dataset.}
    \label{tab:abl}
\end{table}

\section{Conclusion}
This study addresses a key limitation of self-supervised 4D urban scene reconstruction: the unstable dynamic learning of high-motion objects, often caused by discontinuities in input video sequences. To mitigate this issue, we introduce explicit temporal consistency priors derived from a test-time-adapted video generation model.
We further propose a joint timestamp optimization strategy for camera pose alignment, along with an uncertainty distillation strategy for target content extraction. This dual approach effectively balances the distilled temporal prior with the inherent 3D consistency of input views, while simultaneously preventing pixel-level distortions that can arise from generated content.
Our method improves dynamic modeling for fast-moving objects, yielding novel-view renders at test timestamps with reduced distortion and superior visual quality.

\bibliography{aaai2026}

\end{document}